\documentclass{article}

\usepackage[preprint]{neurips_2026}

\usepackage[utf8]{inputenc} 
\usepackage[T1]{fontenc}    
\usepackage{amsfonts}       
\usepackage{nicefrac}       
\usepackage{microtype}      
\usepackage{xcolor}         

\usepackage[utf8]{inputenc} 
\usepackage[T1]{fontenc}    
\usepackage{url}            
\usepackage[colorlinks=true]{hyperref}       
\usepackage{booktabs}       
\usepackage{amsfonts}       
\usepackage{nicefrac}       
\usepackage{microtype}      
\usepackage{xcolor}         
\usepackage{xspace}
\usepackage{soul}

\usepackage{multirow}
\usepackage{booktabs}
\usepackage{xcolor,colortbl}

\usepackage{mwe} 
\usepackage{amsmath}
\usepackage{wrapfig}
\usepackage{floatflt}

\usepackage{algorithm}
\usepackage{algpseudocode}

\newcommand{\duster}{DUSt3R\xspace}

\newcommand{\OurMethodNoSpace}{G3T}
\newcommand{\OurMethod}{\OurMethodNoSpace\xspace}

\newcommand{\OurMethodP}{\OurMethodNoSpace$_P$\xspace}
\newcommand{\OurMethodD}{\OurMethodNoSpace$_D$\xspace}

\newcommand{\simthree}{\mathrm{Sim}(3)}
\newcommand{\simthreeGA}{\mathrm{Sim}_y(3)}

\newcommand{\simthreelie}{\mathfrak{sim}(3)}
\newcommand{\simthreeGAlie}{\mathfrak{sim}_y(3)}

\newcommand{\bharath}[1]{}
\newcommand{\noah}[1]{}

\newcommand{\bestMethod}[1]{\cellcolor{forestgreen!60}\textbf{#1}}
\newcommand{\nextBestMethod}[1]{\cellcolor{forestgreen!25}\underline{#1}}

\definecolor{lightgreen}{RGB}{198, 239, 206}

\definecolor{forestgreen}{RGB}{34, 139, 34}

\title{G3T Up! Gravity Aligned Coordinate Frames Simplify Pointmap Processing}

\author{%
  Bharath Raj Nagoor Kani \hspace{7.5mm} Noah Snavely \vspace{2pt} \\
  Cornell University \vspace{5pt}\\
  {\tt \small~Project Page: \url{https://g3t-paper.github.io/}}
}

\begin{document}

\maketitle

\vspace{-10pt}
\begin{abstract}
 Modern feed-forward 3D reconstruction methods like VGGT predict pixel-aligned pointmaps in camera-centric coordinate frames. However, this choice of coordinate frame is not always optimal. We propose instead to predict pointmaps in upright, gravity-aligned frames that exploit strong structural cues present in many real-world scenes. Unlike camera-centric frames, gravity-aligned frames share a common vertical axis across viewpoints, reducing the rotational degrees of freedom needed to relate pointmaps to one another. To this end, we introduce the Gravity Grounded Geometry Transformer (G3T), fine-tuned from existing models on gravity-aligned 3D data. G3T produces highly accurate gravity-aware predictions, including upright pointmaps and camera-to-gravity poses. We further introduce G3T-Long, a submap-based incremental 3D reconstruction pipeline that leverages the reduced rotational degrees of freedom afforded by upright frames to achieve significantly improved reconstruction accuracy.
\end{abstract}

\section{Introduction}
\label{sec:intro}

Pointmaps have recently become a widespread choice for representing image-centric 3D geometry in learned models. Popularized by \duster~\cite{Wang_2024_CVPR}, they have become a standard representation in systems that perform feedforward 3D reconstruction~\cite{wang2025vggt, wang2025pi3, keetha2026mapanything, depthanything3, Yang_2025_Fast3R}. 

A key design choice for a pointmap representation is \emph{what coordinate frame it is defined in}. The default choice of popular methods is to represent predicted pointmaps in the coordinate frame of the first camera. 
While this is convenient and enables large-scale training on diverse datasets, it is not the only choice, and perhaps not the best choice for all tasks. 

If the end goal is to create a 3D reconstruction of a scene from a set of images, all that matters is that the model can robustly reason about relative transformation between many pointmaps, and hence there is no restriction that we must use \emph{camera} coordinate frames. 

A good coordinate frame for joint 3D alignment across many views might be one that is aligned to strong structural regularities observed in a scene---that is, a world-centric, rather than camera-centric coordinate frame. 
Typical camera-centric 3D reconstruction methods like VGGT~\cite{wang2025vggt} involve a strong model asymmetry where the first image gets to define the coordinate frame, involving whatever arbitrary camera orientation that image might possess (including both camera pitch and roll). 
If instead the set of input images could agree on some natural world-centric coordinate frame, then this asymmetry would be resolved, potentially making for more stable, robust predictions.
Unfortunately, it is difficult to define what a consistent world-centered coordinate frame would be. 

\begin{figure*}
    \centering
    \includegraphics[width=\textwidth]{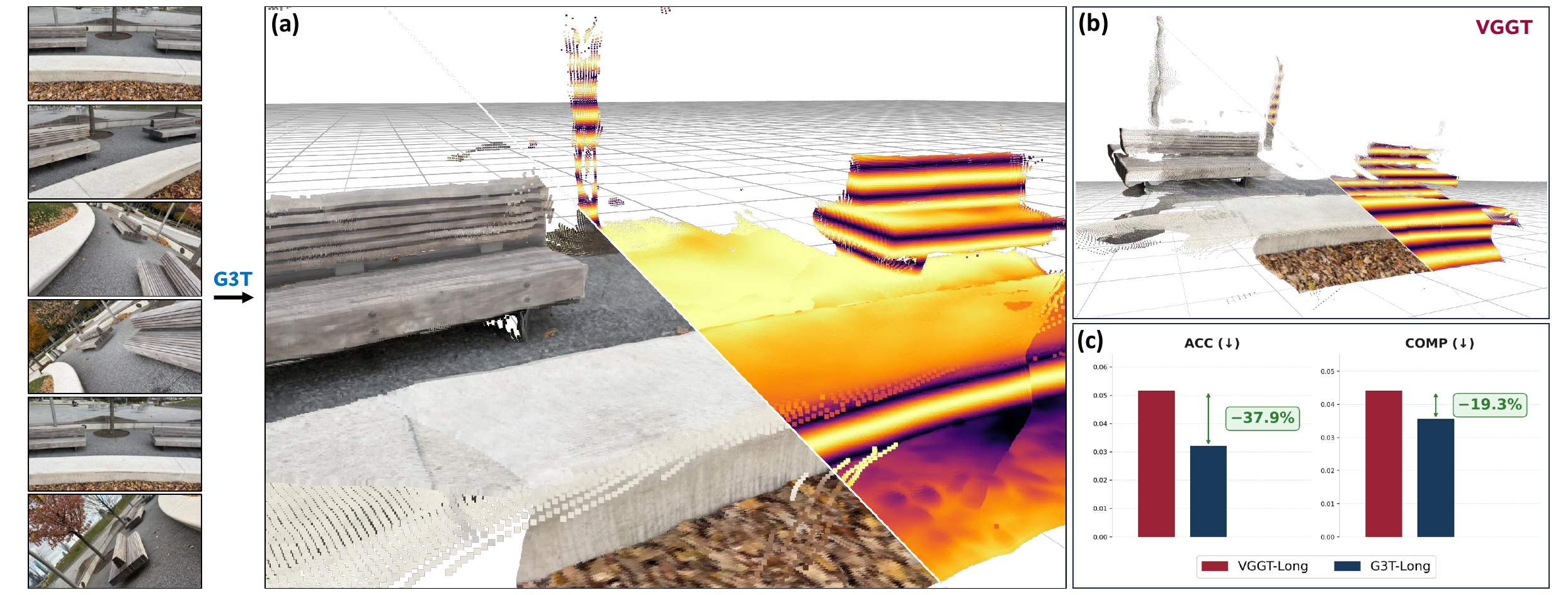}
    \caption{\textbf{Gravity Grounded Geometry Transformer (G3T)} predicts pointmaps aligned with scene gravity, leveraging structural cues inherent in natural scenes. We visualize uprightness using a ground-parallel grid and height-dependent color encoding. Pointmaps produced by \OurMethod show near-constant color in \emph{ground-parallel regions} (such as floors, benches etc.) indicating upright-alignment (a), whereas those produced by VGGT show rapidly varying colors in the same regions (b). Exploiting the reduced rotational degrees of freedom afforded by upright frames, we introduce \OurMethod-Long: a robust, submap-based incremental 3D reconstruction pipeline. Results in (c) report median structure metrics averaged over 10 sequences from the TUM RGBD dataset (see Table~\ref{tab:inc_recon_tum_struc} for full results).}
    \label{fig:teaser}
\end{figure*}

In this work, we propose \textbf{G}ravity \textbf{G}rounded \textbf{G}eometry \textbf{T}ransformer (\textbf{G3T}), a feed-forward 3D prediction model that builds upon VGGT~\cite{wang2025vggt} and predicts pointmaps in \emph{gravity-aligned} coordinate frames, exploiting strong structural regularities induced by gravity (see Figure~\ref{fig:teaser}).

\begin{figure}[t!]
    \centering
    \includegraphics[width=\textwidth]{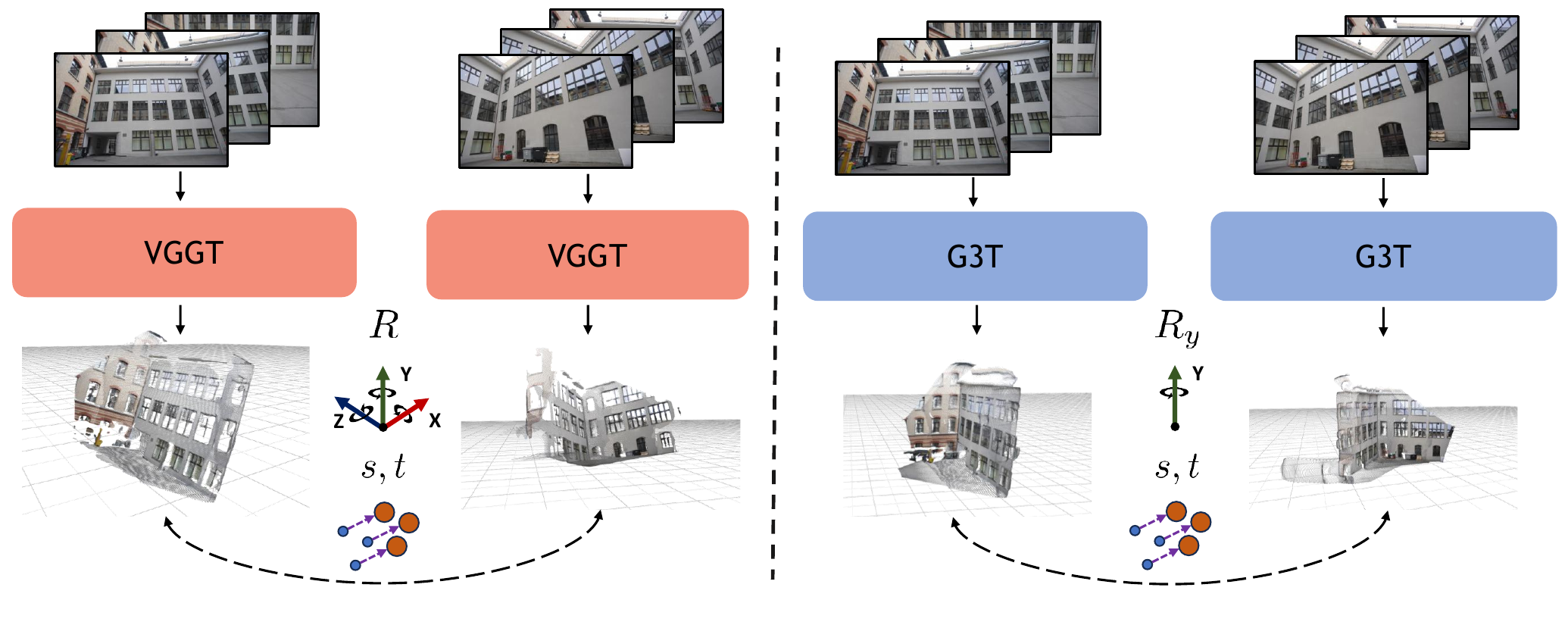}
    \caption{
    \textbf{Camera-aligned and gravity-aligned coordinate frames.} Feedforward 3D reconstruction methods such as VGGT~\cite{wang2025vggt} typically predict pointmaps in a camera coordinate frame. Such pointmaps can be related to each other using a pose $\pi(s, R, t) \in \simthree$ that has 7 degrees of freedom (DoF). In contrast, \OurMethod predicts pointmaps in a gravity-aligned coordinate frame. Such pointmaps can be related to each other using a pose $\pi^y(s, R_{y}, t) \in \simthreeGA$  that has 5-DoF (i.e., rotations are restricted to be rotations along the $y$-axis, which have 1-DoF). We leverage this property to constrain submap-based reconstruction algorithms (as described in Section~\ref{sec:gravity-aligned-procrustes}).
    } 
    \label{fig:grav_coord_frame}
\end{figure}

Pointmaps predicted in such upright frames can be related to each other by a reduced 1 degree of freedom (1-DoF) rotation about the vertical axis (see Figure~\ref{fig:grav_coord_frame}). 
The resulting predictions are natively defined in a natural, upright coordinate frame, making them immediately suitable for downstream tasks such as visualization or simulation without extra post-processing. 
We show that G3T yields camera poses and pointmaps that are more accurately aligned to gravity compared to baseline methods.

We further leverage these structured coordinate frames to develop \textbf{G3T-Long}, a submap-based incremental 3D reconstruction pipeline that extends VGGT-Long~\cite{deng2025vggtlongchunkitloop}. By restricting alignment to the reduced rotational degrees of freedom afforded by upright frames, G3T-Long achieves significantly more accurate reconstructions with reduced drift. While we instantiate these ideas on top of VGGT and VGGT-Long, the concept of using gravity-aligned coordinate frames is general and can be applied to other camera-centric methods as well.

\section{Related work}
\label{sec:related}

\smallskip
\noindent \textbf{Multi-view 3D pointmap prediction.} \duster~\cite{Wang_2024_CVPR} introduced a transformer-based architecture for pixel-aligned pointmap prediction from unposed image pairs. Follow-up work extended it predict in metric space~\cite{leroy2024grounding}, handle dynamic scenes~\cite{zhang2024monst3r, jin2024stereo4d, chen2025easi3r, st4rtrack2025}, and process input in a streaming fashion~\cite{wang20243dspann3r, cut3r, chen2025ttt3r}. More recent work has shifted toward direct feedforward prediction from multi-view images~\cite{wang2025vggt, tang2024mvdust3r, Yang_2025_Fast3R, depthanything3, keetha2026mapanything} with with subsequent work extending these methods to process long sequences~\cite{deng2025vggtlongchunkitloop, maggio2025vggt-slam, zhang2026loger, jin2026zipmap}. Despite these advances, little attention has been paid to the choice of coordinate frame. Of note, while $\pi^3$~\cite{wang2025pi3} achieves invariance to input ordering, predictions remain in an arbitrary frame that does not exploit natural scene priors. Our focus is to explore the use of alternate, more world-centric coordinate frames for pointmap prediction.

\smallskip
\noindent \textbf{Scene-level canonical frames from image level cues.} Images often possess rich geometric cues such as straight lines, vanishing points that indicate dominant scene orientations, and objects that are upright or resting on flat surfaces. Classic work has sought to detect vanishing points from single images (e.g., by grouping detected lines) to aid in geometric calibration~\cite{shufelt1999performance,tardif2009noniterative,caprile1990using,lee2014upright}. More recently, learning-based approaches~\cite{xian2019uprightnet, jin2022PerspectiveFields, veicht2024geocalib} were proposed to estimate gravity direction (and other properties) in a data-driven fashion. Other related work~\cite{wang2019normalized, kulkarni2019csm} has explored \emph{object-centric} coordinate frames, taking advantage of the fact that many object categories (cars, chairs, airplanes, etc) define natural coordinate frames. However, general scenes often do not have a single, unambiguous world coordinate frame---but every scene on Earth has a gravity direction, making gravity-aligned coordinate frames an attractive choice for pointmap prediction tasks.

\smallskip
\noindent \textbf{Leveraging gravity direction as a prior.}
Gravity direction has been leveraged as a prior in numerous vision tasks such as camera parameter estimation~\cite{kukelova2010closed,homography2017saurer}, rotation averaging~\cite{pan2024gravity}, and human motion reconstruction~\cite{shen2024gvhmr}. To the best of our knowledge, leveraging gravity direction in the context of pointmap prediction has not been previously explored.
In fact, our work is motivated in part by the fact that, once a scene is in a more structured coordinate frame like those produced by our method, other tasks such as stereo estimation can benefit from stronger priors (see, for instance, work like Manhattan world stereo~\cite{furukawa2009manhattan} and schematic surface reconstruction~\cite{wu2012schematic}).

\section{Method}
\label{sec:method}

\subsection{Preliminaries: pointmap prediction in camera frames}

\label{sec:preliminary}

Given a set of $N$ input images $\mathcal{I} = \{ I_i \}$, VGGT~\cite{wang2025vggt} predicts: 
1) a set of pixel-aligned pointmaps $\mathcal{X}^{C_1} = \{X^{C_1}_i \}$ and associated confidence maps $\Psi_{\mathcal{X}^{C_1}} = \{ \psi_{X^{C_1}_i}\}$ (where $I_i, X^{C_1}_i \in \mathbb{R}^{H \times W \times 3}$ and $\psi_{X^{C_1}_i} \in \mathbb{R}^{H \times W }$), 
2) a set of depthmaps $\mathcal{D} = \{D_i\}$ and associated confidence maps $\Psi_\mathcal{D} = \{ \psi_{D_i}\}$ (where $D_i, \psi_{D_i} \in \mathbb{R}^{H \times W }$) and 
3) camera parameters ${\mathcal{G} = \{G_i\}}$ (where $G_i \in \mathbb{R}^9$ is the concatenation of a rotation quaternion $q_i \in \mathbb{R}^4$, translation vector $t_i \in \mathbb{R}^3$ and field of view $f_i \in \mathbb{R}^2$). Note that the user has the choice of using either the predicted pointmaps directly, or using pointmaps created by unprojecting the depthmaps using the camera parameters. 

\subsection{Gravity-aligned pointmap prediction} 

\label{sec:gravity_coords}
\noindent \textbf{Coordinate systems.} A pointmap $X^{C}$ in a camera coordinate frame can be related to a pointmap $X^{G}$ in a \emph{gravity-aligned} coordinate frame (that is, an upright frame) through the application of roll and pitch rotations such that the $y$-axis becomes gravity aligned (see Figure \ref{fig:grav_coord_frame}). In other words, $X^G = R_x  R_z  X^C$, where $R_z$ and $R_x$ represent suitable 3D roll and pitch rotations, respectively. 
While there are an infinite number of gravity-aligned coordinate frames, each related by a rotation about the up axis (the $y$-axis, in our case), the application of just roll and pitch (and not yaw) rotations uniquely determines a specific gravity-aligned coordinate frame for a given image (just as each image has its own camera frame). Note that the gravity-aligned coordinate frame has the same origin as that of the camera frame.

\begin{figure*}[t!]
    \centering
    \includegraphics[width=0.8\textwidth]{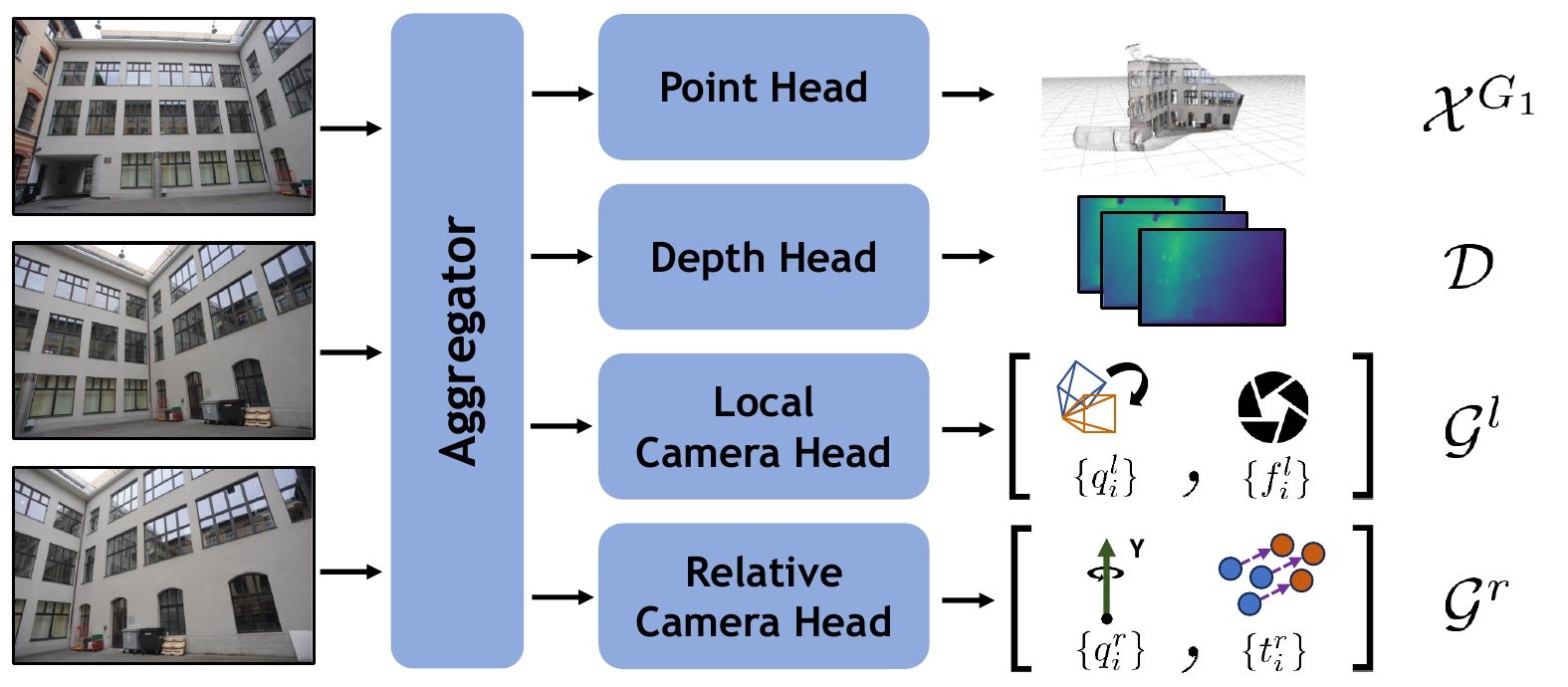}
    \caption{\textbf{Model architecture.} \OurMethod builds upon VGGT with two key modifications. First, the point head outputs pointmaps in the \emph{gravity-aligned} frame of the first image ($\mathcal{X}^{G_1} = \{X^{G_1}_i\}$). Second, we replace VGGT's camera head with two new heads: the \emph{local camera head}, whose outputs capture gravity-to-camera rotation and camera intrinsics parameters in $\mathcal{G}^l = \{G^l_i\}$; and the \emph{relative camera head}, which capture 1-DoF relative yaw and translation parameters in $\mathcal{G}^r = \{G^r_i\}$. The aggregator, depth head, and point head architectures are otherwise unchanged from VGGT. }
    \label{fig:model_arch}
\end{figure*}

\vspace{5pt}
\noindent \textbf{Pointmap prediction in gravity frames.} We propose \textbf{\OurMethod}, which builds upon VGGT with two key modifications (Figure~\ref{fig:model_arch}). First, the point head now outputs pointmaps in the gravity frame of the first image ($G_1$) rather than its camera frame ($C_1$); we denote these as $\mathcal{X}^{G_1} = \{X^{G_1}_i\}$. Second, we replace VGGT's camera head with two new heads: the \emph{local camera head}, which outputs $\mathcal{G}^l = \{G^l_i\}$ where $G^l_i \in \mathbb{R}^6$ concatenates a rotation quaternion $q^l_i \in \mathbb{R}^4$ (gravity-to-camera rotation) and field of view $f^l_i \in \mathbb{R}^2$ (intrinsics); and the \emph{relative camera head}, which outputs $\mathcal{G}^r = \{G^r_i\}$ where $G^r_i \in \mathbb{R}^5$ concatenates rotation parameters $q^r_i \in \mathbb{R}^2$ (1-DoF relative yaw w.r.t.\ the first frame) and translation $t^r_i \in \mathbb{R}^3$ (relative translation w.r.t.\ the first frame). The aggregator, depth head, and point head architectures are otherwise unchanged from VGGT. As in VGGT, gravity-aligned pointmaps can also be obtained by unprojecting depthmaps by using the composed local and relative poses.

\subsection{Constraining submap-based incremental 3D reconstruction}

\smallskip
\noindent \textbf{Incremental 3D reconstruction via submap alignment.} While VGGT can process an arbitrary number of images, in practice, the user is limited by GPU memory constraints. 
To this end, VGGT-Long~\cite{deng2025vggtlongchunkitloop} proposes splitting a sequence of images into overlapping chunks in a sliding window fashion and then aligning the chunk-wise pointmap predictions. 

Suppose $\mathcal{A}^{C_1} = \{X^{C_1}_i \}$ and $\mathcal{B}^{C_2} = \{X^{C_2}_j \}$ are pointmap predictions obtained from two chunks with some overlapping images. Let $\mathcal{M}$ be the set of indices $(i,j)$ that refer to the same image but in different chunks. Then, we can denote the set of corresponding pointmaps as $\{ (X^{C_1}_{i}, X^{C_2}_{j}) \}$, where $(i, j) \in \mathcal{M}$. Given this, the relative $\simthree$ transformation $\pi$ that aligns $\mathcal{B}^{C_2}$ to $\mathcal{A}^{C_1}$ can be computed using the following least squares objective:
\begin{align}
    \pi^* = \arg\min_{\pi \in \simthree} \sum_{(i,j) \in \mathcal{M}} \| X_{i}^{C_1} - \pi X_{j}^{C_2} \|^2_2 \label{eq:chunk_align}
\end{align}
This objective can be solved in closed form using Procrustes alignment~\cite{umeyamaProcrustes}. 
In practice, VGGT-Long uses an iteratively re-weighted version of Eq.~\ref{eq:chunk_align} (where the weights are initialized using confidence maps corresponding to the pointmaps; the interested reader can refer to their paper~\cite{deng2025vggtlongchunkitloop} for details). The relative transformations are computed between two types of chunks: 
1) adjacent overlapping chunks, and
2) chunks constructed across loop closure detections. 
Once these are computed, VGGT-Long performs a global optimization over the absolute $\simthree$ transformations of all chunks to be consistent with these relative measurements. 
The optimization objective is (roughly) given by:
\begin{align}
    \{\pi_k^*\} = \arg\min_{\{\pi_k\}} & \sum_{k=1}^{K-1} \|  \delta_\pi(k, k+1) \|_2^2 + \sum_{(i,j) \in \mathcal{L}} \| \delta_\pi(i, j) \|_2^2 \label{eq:lm_opt}
\end{align}
Here, $\delta_\pi(i, j) = {\tilde{\pi}}_{i,j}^{-1} \pi_i^{-1} \pi_j$ measures the discrepancy between the relative pose ${\tilde{\pi}}_{i,j}$ estimated during chunk alignment and the relative pose derived by composing the absolute poses $\pi_i, \pi_j$ being optimized. $K$ denotes the number of overlapping chunks and $\mathcal{L}$ denotes the set of loop closure chunks. The norms in Equation~\ref{eq:lm_opt} are applied to the Lie algebra representations of $\delta_\pi$, as optimization is performed over $\simthreelie$ (the 7-dimensional Lie algebra of $\simthree$) allowing the problem to be solved in an unconstrained manner using the Levenberg-Marquardt algorithm (see supplementary for more details)

\label{sec:gravity-aligned-procrustes}

\begin{figure}[t]
\centering
\scalebox{0.95}{
\begin{minipage}[t][5.2cm][t]{0.49\textwidth}
\begin{algorithm}[H]
\caption{\emph{Procrustes:} Estimate $\pi(s, R, t)$ from pointmaps $\mathbf{P}$, $\mathbf{Q}$}
\label{alg:estimate_sim3}
\begin{algorithmic}[1]

\State $\tilde{\mathbf{P}}, \tilde{\mathbf{Q}} \leftarrow \mathbf{P} - \boldsymbol{\mu}_\mathbf{P}, \mathbf{Q} - \boldsymbol{\mu}_\mathbf{Q}$ 

\State $s \leftarrow \|\tilde{\mathbf{Q}}\|_{\text{rms}} \;/\; \|\tilde{\mathbf{P}}\|_{\text{rms}}$

\Statex
\vspace{10pt}

\State $\mathbf{H} \leftarrow (s\tilde{\mathbf{P}})^\top \tilde{\mathbf{Q}}$

\State $\mathbf{U}, \boldsymbol{\Sigma}, \mathbf{V}^\top \leftarrow \text{SVD}(\mathbf{H})$ 
\State $\mathbf{R} \leftarrow \mathbf{V} \mathbf{U}^\top$

\Statex
\vspace{6pt}

\State $\mathbf{t} \leftarrow \boldsymbol{\mu}_\mathbf{Q} - s\,\mathbf{R}\,\boldsymbol{\mu}_\mathbf{P}$

\State \textbf{return} $s,\, \mathbf{R},\, \mathbf{t}$

\end{algorithmic}
\end{algorithm}
\vspace{\fill}
\end{minipage}
\hfill
\begin{minipage}[t][5.2cm][t]{0.49\textwidth}
\begin{algorithm}[H]
\caption{\emph{GA-Procrustes:} Estimate $\pi^y(s, R_y, t)$ from pointmaps $\mathbf{P}$, $\mathbf{Q}$}
\label{alg:estimate_sim3GA}
\begin{algorithmic}[1]

\State $\tilde{\mathbf{P}}, \tilde{\mathbf{Q}} \leftarrow \mathbf{P} - \boldsymbol{\mu}_\mathbf{P}, \mathbf{Q} - \boldsymbol{\mu}_\mathbf{Q}$ 

\State $s \leftarrow \|\tilde{\mathbf{Q}}\|_{\text{rms}} \;/\; \|\tilde{\mathbf{P}}\|_{\text{rms}}$

\State \colorbox{lightgreen}{$\tilde{\mathbf{P'}}, \tilde{\mathbf{Q'}} \leftarrow \text{project}_{xz}(\tilde{\mathbf{P}}), \text{project}_{xz}(\tilde{\mathbf{Q}})$} 

\State $\mathbf{H'} \leftarrow (s\tilde{\mathbf{P'}})^\top \tilde{\mathbf{Q'}}$

\State $\mathbf{U'}, \boldsymbol{\Sigma'}, \mathbf{V'}^\top \leftarrow \text{SVD}(\mathbf{H'})$ 
\State $\mathbf{R'} \leftarrow \mathbf{V'} \mathbf{U'}^\top$

\State \colorbox{lightgreen}{$\mathbf{R_y} \leftarrow \text{lift}_{xyz}(\mathbf{R'})$}

\State $\mathbf{t} \leftarrow \boldsymbol{\mu}_\mathbf{Q} - s\,\mathbf{R_y}\,\boldsymbol{\mu}_\mathbf{P}$

\State \textbf{return} $s,\, \mathbf{R_y},\, \mathbf{t}$

\end{algorithmic}
\end{algorithm}
\end{minipage}
}
\end{figure}

\smallskip
\noindent \textbf{Gravity-aligned Procrustes alignment.} Given two pointmaps $X_i^{C_i}, X_i^{C_j}$ derived from the same image but represented in different coordinate frames $C_i$ and $C_j$, we can estimate their relative pose using Procrustes~\cite{umeyamaProcrustes} (Algorithm~\ref{alg:estimate_sim3}, where $\mathbf{P}$, $\mathbf{Q}$ are corresponding pointmaps, $\boldsymbol{\mu}_\mathbf{P}, \boldsymbol{\mu}_\mathbf{Q}$ their means, and $\|.\|_{\text{rms}}$ the root mean squared distance). This yields a 7-DoF pose $\pi(s, R, t) \in \simthree$, with 3D rotation $R$, 3D translation $t$, and 1D scale $s$. We write $\pi = \mathcal{P}(X_i^{C_i}, X_i^{C_j})$ to denote the application of Procrustes.

Since gravity-aligned pointmaps are related by a 1-DoF rotation about the $y$-axis, we constrain Procrustes to a 2D rotation on the $xz$-plane, yielding our gravity aligned, i.e., \emph{GA-Procrustes} algorithm (Algorithm~\ref{alg:estimate_sim3GA}, where $\text{project}_{xz}$ and $\text{lift}_{xyz}$ project to the $xz$-plane and reparameterize the result as a 3D $y$-axis rotation, respectively). GA-Procrustes produces a 5-DoF pose $\pi_{y}(s, R_{y}, t) \in \simthreeGA$, where $\simthreeGA \subset \simthree$ restricts rotations to the $y$-axis. We write $\pi_y = \mathcal{P}_{\text{GA}}(X_i^{G_i}, X_i^{G_j})$ to denote the application of GA-Procrustes.

\vspace{5pt}
\noindent \textbf{Gravity-aligned incremental 3D reconstruction.} We now describe how to construct \OurMethod-Long, a gravity-aligned variant of VGGT-Long. The two key modifications are as follows. First, we replace Procrustes with GA-Procrustes in the chunk alignment procedure, which naturally constrains the estimated relative poses to $\simthreeGA$. This gives the following modified chunk-alignment objective:
\begin{align}
    \pi^{y*} = \arg\min_{\pi^y \in \simthreeGA} \sum_{(i,j) \in \mathcal{M}} \| X_{i}^{G_1} - \pi^y X_{j}^{G_2} \|^2_2 \label{eq:chunk_align_1dof}
\end{align}
Next, we redefine the global optimization over $\simthreeGA$ rather than $\simthree$, with optimization performed over $\simthreeGAlie$ (the 5-dimensional Lie algebra of $\simthreeGA$, reduced from the 7-dimensional $\simthreelie$; see supplementary for more details). This gives us the following modified objective:
\begin{equation}
     \{\pi_k^{y*}\} = \arg\min_{\{\pi_k^{y*}\}}  \sum_{k=1}^{K-1} \|  (\delta_{\pi^y}(k, k{+}1)) \|_2^2 + \negthickspace \sum_{(i,j) \in \mathcal{L}} \negthickspace \|  (\delta_{\pi^y}(i, j)) \|_2^2 \label{eq:lm_opt_1dof}
\end{equation}
All other implementation details remain unchanged.

\begin{figure*}
    \centering
    \includegraphics[width=\textwidth]{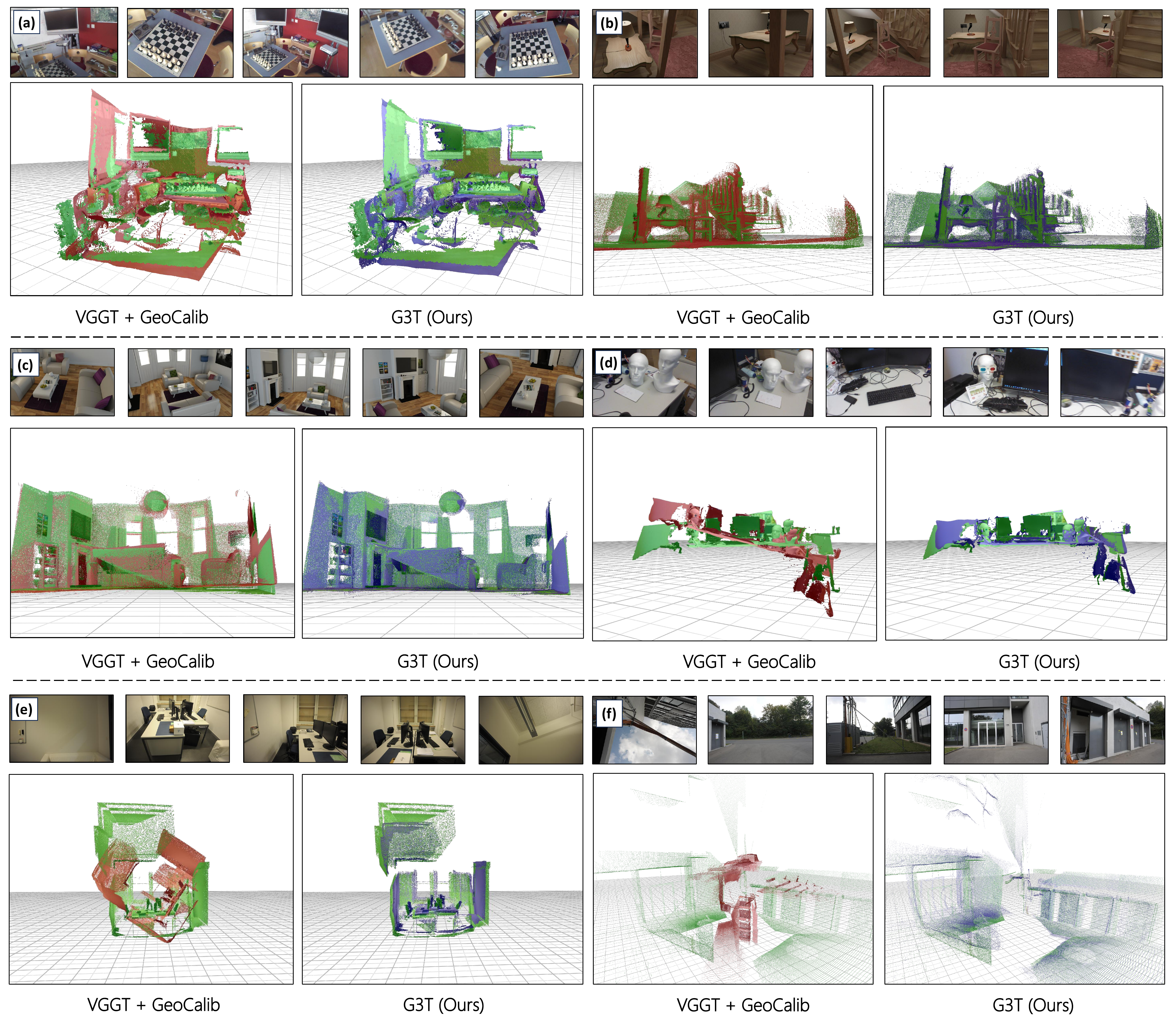}
    \caption{\textbf{\OurMethod can consistently place pointmaps in an upright frame.} For the given set of input images, we compare VGGT predictions (\textbf{\color{red}{red}}) made upright using GeoCalib, and our \OurMethod predictions (\textbf{\color{blue}{blue}}) with ground-truth gravity-aligned pointmaps (\textbf{\color{forestgreen}{green}}). We also render a grid depicting a plane parallel to the ground. We observe that the composition of VGGT and GeoCalib often produces ``slanted'' pointmap predictions that are misaligned with the vertical direction. In contrast, \OurMethod produces pointmaps that are better aligned along the vertical direction, even in scenes captured in images with large pitch angles.} 
    \label{fig:mv_recon}
    \vspace{-10pt}
\end{figure*}

\vspace{-5pt}
\subsection{Training details}

\label{sec:training_details}

To create \OurMethod, we fine-tune the open-source VGGT model on gravity-aligned pointmaps. However, the ground-truth information available found in popular 3D datasets are not necessarily gravity-aligned. 
Inspired by prior work~\citep{tung2024megascenes,veicht2024geocalib} that uses COLMAP's \texttt{model\_orientation\_aligner}~\cite{schoenberger2016mvs, schoenberger2016sfm} for Manhattan alignment, we apply it to gravity-align ground-truth for datasets that are not naturally gravity-aligned (see supplementary for details).
Overall, we use gravity-aligned data from five large-scale datasets (MegaDepth~\cite{megadepth18}, Hypersim~\cite{hypersim21}, ARKitScenes~\cite{dehghan2021arkitscenes}, DL3DV~\cite{ling2024dl3dv} and TartanAir~\cite{tartanair2020iros}) for fine-tuning.

We opt for fine-tuning the entire VGGT architecture (no weights are frozen). We use the same set of losses as VGGT for training, but with gravity-aligned ground truth points. We ignore the track head. Within each batch, we normalize predicted and ground-truth pointmaps independently before computing the loss (as done in \duster~\cite{Wang_2024_CVPR}). Our final model was fine-tuned on 8 A100 GPUs for 40 epochs, with 1000 data samples per epoch (which took approximately 1 week). For each batch, we randomly sample 2-12 views per scene for a maximum of 96 images per GPU. We use a gradient accumulation of 4 and a cosine learning rate schedule that starts at 1e-6 after 1 warmup epoch. The remaining training configuration mimics the released version of VGGT.

\section{Experiments}
\label{sec:expts}

We begin our analysis by first evaluating the uprightness of the camera-to-gravity poses and pointmaps predicted by \OurMethod (Section \ref{sec:c2g_expt}). Then, we analyze the effectiveness of performing gravity-aligned incremental 3D reconstruction using \OurMethod (Section \ref{sec:incremental_3D_expt}).

\subsection{Evaluating uprightness}

\label{sec:c2g_expt}

\smallskip
\noindent \textbf{Evaluation setup.} To evaluate the uprightness of \OurMethod's predictions, we estimate the camera-to-gravity rotation via two methods: (1) \textit{Local Head}, which directly extracts the rotation from local camera head outputs $\mathcal{G}^l$, and (2) \textit{Procrustes}, where Procrustes is used on the pointmap outputs of \OurMethod and VGGT. 
As a baseline, we evaluate GeoCalib~\cite{veicht2024geocalib} (with multi-image optimization). Given that GeoCalib can be applied as a post-process to make VGGT upright, we are effectively testing this composed strategy against direct gravity-frame prediction by \OurMethod. 
To assess the impact of fine-tuning on structure quality, we additionally compute accuracy (ACC), completeness (COMP), and normal consistency (NC) following prior work~\cite{cut3r, wang20243dspann3r, wang2025vggt, wang2025pi3}, after aligning predicted and ground-truth pointmaps via Procrustes. 
Pointmaps are obtained either directly from model predictions (VGGT$_P$ and \OurMethodP) or by unprojecting depth predictions using estimated cameras (VGGT$_D$ and \OurMethodD).

For camera-to-gravity evaluation, we test with 1, 4, and 8 input views on 7Scenes~\cite{shotton2013scene}, NRGBD~\cite{Azinovic_2022_CVPR}, and ETH3D~\cite{schoeps2017cvpr}, all of which are unseen during training for both models. For structure evaluation, we use sets of five input views from the same datasets. We elaborate on the pre-processing and view sampling setup in the supplementary.

\smallskip
\noindent \textbf{On using Procrustes.} Given a set of $N$ images $\mathcal{I}$, suppose we predict a set of gravity-aligned pointmaps $\mathcal{X}^{G_1}$ using \OurMethod, and a set of corresponding camera-frame pointmaps $\mathcal{X}^{C_1}$ predicted using VGGT. We can then compute the camera-to-gravity rotation using Procrustes alignment between the two sets of pointmaps. Since pointmap predictions can be noisy, we implement a RANSAC loop which incorporates pointmap confidences to obtain a robust estimate of the rotation matrix (please refer to the supplementary for more details).

\begin{table}[t!]
\centering
\caption{
\textbf{Camera to gravity rotation estimation results.} We evaluate the quality of camera-to-gravity pose estimation when provided with 1, 4, or 8 input images on three datasets that were unseen during training. 
We observe that methods based on \OurMethod (in particular Local Head) performs much better than GeoCalib across most metrics. Additionally, methods based on \OurMethod show greater reduction in mean rotation error than GeoCalib. Rotation errors are given in degrees.}
\label{tab:c2g}
\resizebox{\textwidth}{!}{%
\begin{tabular}{llccccccccccccccc}
\toprule
\multirow{3}{*}{Dataset} & \multirow{3}{*}{Method} & \multicolumn{5}{c}{1V} & \multicolumn{5}{c}{4V} & \multicolumn{5}{c}{8V} \\
\cmidrule(lr){3-7} \cmidrule(lr){8-12} \cmidrule(lr){13-17}
 & & \multicolumn{2}{c}{$R_{\text{err}}$ ($\downarrow$)} & \multicolumn{3}{c}{$R_{\text{acc}}$ ($\uparrow$)} & \multicolumn{2}{c}{$R_{\text{err}}$ ($\downarrow$)} & \multicolumn{3}{c}{$R_{\text{acc}}$ ($\uparrow$)} & \multicolumn{2}{c}{$R_{\text{err}}$ ($\downarrow$)} & \multicolumn{3}{c}{$R_{\text{acc}}$ ($\uparrow$)} \\
\cmidrule(lr){3-4} \cmidrule(lr){5-7} \cmidrule(lr){8-9} \cmidrule(lr){10-12} \cmidrule(lr){13-14} \cmidrule(lr){15-17}
 & & Mean & Median & @1\textdegree & @2\textdegree & @5\textdegree & Mean & Median & @1\textdegree & @2\textdegree & @5\textdegree & Mean & Median & @1\textdegree & @2\textdegree & @5\textdegree \\
\midrule
\multirow{3}{*}{7Scenes} & GeoCalib & 6.78 & 3.07 & 9.60 & 29.20 & 74.80 & 6.60 & 2.79 & 11.20 & 32.00 & 76.80 & 6.56 & 2.75 & 10.80 & 32.40 & 77.60 \\
& Procrustes & \nextBestMethod{2.00} & \nextBestMethod{1.73} & \nextBestMethod{15.60} & \nextBestMethod{60.40} & \nextBestMethod{96.40} & \nextBestMethod{1.87} & \nextBestMethod{1.59} & \nextBestMethod{23.60} & \nextBestMethod{63.60} & \nextBestMethod{97.20} & \nextBestMethod{1.88} & \nextBestMethod{1.64} & \nextBestMethod{22.00} & \nextBestMethod{61.60} & \nextBestMethod{97.20} \\
& Local Head & \bestMethod{1.92} & \bestMethod{1.68} & \bestMethod{18.00} & \bestMethod{64.40} & \bestMethod{96.80} & \bestMethod{1.78} & \bestMethod{1.54} & \bestMethod{26.00} & \bestMethod{66.80} & \bestMethod{98.00} & \bestMethod{1.78} & \bestMethod{1.49} & \bestMethod{26.00} & \bestMethod{65.20} & \bestMethod{98.40} \\
\midrule
\multirow{3}{*}{NRGBD} & GeoCalib & 2.61 & 1.91 & 26.80 & 53.20 & 89.60 & 2.28 & 1.50 & 38.80 & 65.20 & 92.00 & 2.19 & 1.30 & 41.20 & 68.40 & 91.60 \\
& Procrustes & \nextBestMethod{1.47} & \nextBestMethod{0.71} & \nextBestMethod{60.40} & \nextBestMethod{78.40} & \nextBestMethod{93.60} & \nextBestMethod{1.32} & \nextBestMethod{0.60} & \nextBestMethod{70.00} & \nextBestMethod{78.00} & \bestMethod{95.60} & \nextBestMethod{1.26} & \nextBestMethod{0.50} & \nextBestMethod{71.20} & \nextBestMethod{78.80} & \bestMethod{95.20} \\
& Local Head & \bestMethod{1.33} & \bestMethod{0.63} & \bestMethod{65.60} & \bestMethod{78.80} & \bestMethod{94.00} & \bestMethod{1.21} & \bestMethod{0.46} & \bestMethod{76.80} & \bestMethod{80.40} & \nextBestMethod{95.20} & \bestMethod{1.13} & \bestMethod{0.40} & \bestMethod{79.20} & \bestMethod{80.40} & \nextBestMethod{94.00} \\
\midrule
\multirow{3}{*}{ETH3D} & GeoCalib & 2.24 & \nextBestMethod{0.73} & \nextBestMethod{68.80} & \nextBestMethod{90.40} & 97.20 & 2.22 & \nextBestMethod{0.71} & \nextBestMethod{69.20} & \nextBestMethod{91.20} & \nextBestMethod{97.60} & 2.21 & \nextBestMethod{0.72} & \nextBestMethod{69.60} & \nextBestMethod{90.80} & \nextBestMethod{98.00} \\
& Procrustes & \nextBestMethod{1.96} & 1.08 & 44.00 & 78.80 & \nextBestMethod{98.00} & \nextBestMethod{1.95} & 1.00 & 50.00 & 80.40 & 95.60 & \nextBestMethod{1.85} & 1.03 & 47.60 & 80.40 & 96.00 \\
& Local Head & \bestMethod{1.62} & \bestMethod{0.60} & \bestMethod{79.20} & \bestMethod{95.60} & \bestMethod{98.40} & \bestMethod{1.11} & \bestMethod{0.55} & \bestMethod{81.20} & \bestMethod{97.60} & \bestMethod{98.80} & \bestMethod{1.05} & \bestMethod{0.57} & \bestMethod{81.60} & \bestMethod{98.00} & \bestMethod{98.80} \\
\bottomrule
\end{tabular}%
}
\end{table}

\begin{table}[t!]
\centering
\caption{
\textbf{Multi-view pointmap errors.} We compare the performance of VGGT and \OurMethod on structure metrics on three datasets that were unseen during training. We observe that \OurMethod retains similar structure performance while having the property of being upright (as observed in Table~\ref{tab:c2g}). Furthermore, the depthmap unprojection results indicate that the relative yaw predictions output by \OurMethod are stable as well.
}
\label{tab:mv_table_frame_indep}
\resizebox{0.9\linewidth}{!}{%
\setlength{\tabcolsep}{6pt}
\begin{tabular}{l ccc ccc ccc}
\toprule
& \multicolumn{3}{c}{7Scenes} & \multicolumn{3}{c}{NRGBD} & \multicolumn{3}{c}{ETH3D} \\
\cmidrule(lr){2-4} \cmidrule(lr){5-7} \cmidrule(lr){8-10}
Model & ACC ($\downarrow$) & COMP ($\downarrow$) & NC ($\uparrow$) & ACC ($\downarrow$) & COMP ($\downarrow$) & NC ($\uparrow$) & ACC ($\downarrow$) & COMP ($\downarrow$) & NC ($\uparrow$) \\
\midrule
VGGT$_P$    & 0.029                  & 0.034                  & \bestMethod{0.796}     & \nextBestMethod{0.024} & \nextBestMethod{0.019} & \bestMethod{0.921}     & \nextBestMethod{0.191} & 0.191                  & \nextBestMethod{0.890} \\
VGGT$_D$    & 0.031                  & \nextBestMethod{0.032} & 0.753                  & \bestMethod{0.022}     & \bestMethod{0.018}     & \nextBestMethod{0.913} & 0.209                  & \nextBestMethod{0.174} & 0.880 \\
\OurMethodP & \nextBestMethod{0.028} & \nextBestMethod{0.032} & \nextBestMethod{0.793} & 0.026                  & 0.021                  & 0.907                  & \bestMethod{0.188}     & 0.181                  & \bestMethod{0.892} \\
\OurMethodD & \bestMethod{0.026}     & \bestMethod{0.029}     & 0.780                  & 0.026                  & 0.021                  & 0.900                  & 0.194                  & \bestMethod{0.165}     & 0.882 \\
\bottomrule
\end{tabular}%
}
\end{table}

\smallskip
\noindent \textbf{Performance measures.} We compare the estimated camera-to-gravity pose of the 1st frame to the corresponding ground-truth. While both the local head of \OurMethod and GeoCalib can provide estimates for all frames, we only evaluate the 1st frame pose: 1) to be consistent with the Procrustes method which outputs the pose estimate only for the 1st frame, and 2) to cleanly measure the impact of adding more inputs views on resolving ambiguities in estimating gravity direction. We use the geodesic rotation error (as defined in~\cite{InterPose}) to measure performance:
\begin{align}
    R_{\text{err}}= \text{arccos} \left ( \frac{\text{Trace}(R_2R_1^T) - 1}{2} \right ) \label{eq:rot_error}
\end{align}
We also compute rotation accuracy by calculating the percentage of samples with rotation error less than 1/2/5 degrees.

\smallskip
\noindent \textbf{Observation 1:} \textit{\OurMethod can estimate high-quality camera-to-gravity poses.} Quantitative results are shown in Table \ref{tab:c2g}. 
We see that the rotation estimates obtained using \OurMethod consistently outperform GeoCalib on most metrics. 
In particular, the mean error for 7Scenes and NRGBD is reduced by more than half. 
Also, we note that the rotation matrix extracted directly from the local head tends to be better than Procrustes. 
Finally, we note that methods based on \OurMethod tends to have a greater reduction in rotation error when the number of views is increased (for example, see the mean rotation error for ETH3D and 7Scenes), thus pointing to the effectiveness of \OurMethod incorporating structural cues from multi-view data.

\begin{table}[t!]
\centering
\caption{
\textbf{Submap-based incremental 3D reconstruction on TUM RGBD: Ambient frame pose metrics.}
We measure the quality of the optimized absolute poses per chunk in each model's \emph{ambient} frame: camera frame for VGGT and gravity frame for \OurMethod (as described in Section \ref{sec:incremental_3D_expt}). $N$ denotes the number of overlapping chunks a sequence was split into. We can see that \OurMethod generally has superior performance in APE metrics. This indicates that leveraging gravity-aligned pointmap predictions and performing chunk alignment and optimization in $\simthreeGA$ improves robustness. Additionally, we observe lesser drift along the $y$-axis when using \OurMethod.
}
\label{tab:inc_recon_tum_pose}
\resizebox{\linewidth}{!}{%
\setlength{\tabcolsep}{4pt}
\begin{tabular}{llcccccccccc}
\toprule
\multirow{2}{*}{Method} & \multirow{2}{*}{Metric ($\downarrow$)} & fr1/360 & fr1/desk & fr1/desk2 & fr1/room & fr1/plant & fr1/teddy & fr3/loh\footnotemark & fr2/ps\footnotemark[\value{footnote}] & fr2/ps2\footnotemark[\value{footnote}] & fr2/ps3\footnotemark[\value{footnote}] \\
 & & $(N\!=\!8)$ & $(N\!=\!6)$ & $(N\!=\!7)$ & $(N\!=\!15)$ & $(N\!=\!13)$ & $(N\!=\!16)$ & $(N\!=\!28)$ & $(N\!=\!24)$ & $(N\!=\!18)$ & $(N\!=\!25)$ \\
\midrule
\multirow{3}{*}{VGGT-Long} & APE$_R$ & \bestMethod{16.320} & 2.313 & \bestMethod{1.553} & 4.276 & 2.963 & 4.732 & 4.289 & 5.924 & 13.845 & 15.228 \\
& APE$_t$ & 0.050 & 0.025 & 0.037 & 0.179 & 0.053 & 0.083 & 0.326 & 0.444 & 0.553 & 0.947 \\
& $\delta_y$ & 0.018 & \bestMethod{0.005} & \bestMethod{0.009} & \bestMethod{0.029} & 0.018 & 0.045 & 0.160 & 0.224 & 0.358 & 0.368 \\
\midrule
\multirow{3}{*}{\OurMethod-Long} & APE$_R$ & 19.309 & \bestMethod{1.434} & 5.192 & \bestMethod{3.502} & \bestMethod{1.429} & \bestMethod{1.948} & \bestMethod{3.964} & \bestMethod{3.482} & \bestMethod{3.512} & \bestMethod{6.381} \\
& APE$_t$ & \bestMethod{0.056} & \bestMethod{0.012} & \bestMethod{0.031} & \bestMethod{0.178} & \bestMethod{0.036} & \bestMethod{0.056} & \bestMethod{0.170} & \bestMethod{0.255} & \bestMethod{0.235} & \bestMethod{0.220} \\
& $\delta_y$ & \bestMethod{0.009} & 0.008 & 0.015 & 0.033 & \bestMethod{0.016} & \bestMethod{0.016} & \bestMethod{0.023} & \bestMethod{0.032} & \bestMethod{0.032} & \bestMethod{0.029} \\
\bottomrule
\end{tabular}%
}
\end{table}

\begin{table}[t!]
\centering
\caption{
\textbf{Submap-based incremental 3D reconstruction on TUM RGBD: Structure metrics.} We measure frame-independent structure metrics to evaluate the reconstruction quality (as described in Section \ref{sec:incremental_3D_expt}). $N$ denotes the number of overlapping chunks a sequence was split into. We can see that \OurMethod generally outperforms VGGT in most cases. These results complement those observed in Table \ref{tab:inc_recon_tum_pose}, as they are a consequence of superior chunk alignment achieved in the gravity frame.
}
\label{tab:inc_recon_tum_struc}
\resizebox{\linewidth}{!}{%
\setlength{\tabcolsep}{4pt}
\begin{tabular}{llcccccccccc}
\toprule
\multirow{2}{*}{Method} & \multirow{2}{*}{Metric ($\downarrow$)} & fr1/360 & fr1/desk & fr1/desk2 & fr1/room & fr1/plant & fr1/teddy & fr3/loh\footnotemark[\value{footnote}] & fr2/ps\footnotemark[\value{footnote}] & fr2/ps2\footnotemark[\value{footnote}] & fr2/ps3\footnotemark[\value{footnote}] \\
 & & $(N\!=\!8)$ & $(N\!=\!6)$ & $(N\!=\!7)$ & $(N\!=\!15)$ & $(N\!=\!13)$ & $(N\!=\!16)$ & $(N\!=\!28)$ & $(N\!=\!24)$ & $(N\!=\!18)$ & $(N\!=\!25)$ \\
\midrule
\multirow{3}{*}{VGGT-Long} & ACC ($\downarrow$) & 0.057 & \bestMethod{0.009} & 0.014 & 0.046 & 0.032 & 0.028 & 0.064 & 0.111 & 0.196 & 0.272 \\
& COMP ($\downarrow$) & 0.071 & 0.012 & 0.016 & 0.043 & 0.041 & 0.044 & 0.044 & 0.048 & 0.140 & 0.201 \\
& NC ($\uparrow$) & 0.701 & \bestMethod{0.628} & \bestMethod{0.619} & \bestMethod{0.664} & 0.619 & 0.594 & 0.606 & 0.663 & 0.700 & 0.678 \\
\midrule
\multirow{3}{*}{\OurMethod-Long} & ACC ($\downarrow$) & \bestMethod{0.043} & \bestMethod{0.009} & \bestMethod{0.013} & \bestMethod{0.040} & \bestMethod{0.024} & \bestMethod{0.021} & \bestMethod{0.024} & \bestMethod{0.059} & \bestMethod{0.069} & \bestMethod{0.084} \\
& COMP ($\downarrow$) & \bestMethod{0.058} & \bestMethod{0.011} & \bestMethod{0.015} & \bestMethod{0.035} & \bestMethod{0.033} & \bestMethod{0.036} & \bestMethod{0.021} & \bestMethod{0.036} & \bestMethod{0.067} & \bestMethod{0.068} \\
& NC ($\uparrow$) & \bestMethod{0.714} & 0.616 & 0.606 & 0.662 & \bestMethod{0.621} & \bestMethod{0.602} & \bestMethod{0.647} & \bestMethod{0.705} & \bestMethod{0.731} & \bestMethod{0.748} \\
\bottomrule
\end{tabular}%
}
\end{table}

\smallskip
\noindent \textbf{Observation 2:} \textit{Pointmaps predicted by \OurMethod have stronger gravity-alignment while retaining structure quality.} The Procrustes results in Table~\ref{tab:c2g} are consistently lower than the corresponding GeoCalib values. 
This provides an indirect measure of pointmap uprightness. This improvement is supported by qualitative visualizations in Figure~\ref{fig:mv_recon}, where we can observe that \OurMethod produces pointmaps with stronger gravity-alignment than those obtained by lifting VGGT predictions to the gravity frame using GeoCalib. 
Additionally, the comparable pointmap structure performance to VGGT in Table~\ref{tab:mv_table_frame_indep} confirms that upright alignment is achieved while retaining structure quality. 
The depthmap unprojection results additionally validate that the relative yaws predicted by the global head are informative.

\footnotetext{We shorten \texttt{long\_office\_household} to \texttt{loh} and \texttt{pioneer\_slam} to \texttt{ps} for brevity.}

\subsection{Submap-based incremental 3D reconstruction}

\label{sec:incremental_3D_expt}

\smallskip
\noindent \textbf{Evaluation setup.} We evaluate two quantities to measure the effectiveness of gravity-aligned submap-based incremental reconstruction. First, we measure the quality of the optimized absolute poses per chunk against ground-truth poses in their respective \emph{ambient} frame. In other words, the ground-truth poses are camera-to-world poses for VGGT (which operates in $\simthree$) and are gravity-to-world poses for \OurMethod (which operates in $\simthreeGA$). Second, we measure overall structure quality after incremental reconstruction. For both experiments, we use pointmaps obtained from the point head.

We evaluate on sequences selected from the TUM RGBD dataset~\cite{sturm12benchmark}. As described in Section~\ref{sec:gravity-aligned-procrustes}, sequences are broken into overlapping chunks are then processed in a sliding window fashion. We set the maximum chunk size to 25 frames, with 7 overlapping frames between adjacent chunks. Further details on preprocessing, hyperparameters, and analyses (including the impact of different hyperparameter choices and usage of depthmap-unprojected points) are provided in the supplementary.

\smallskip
\noindent \textbf{Performance measures.} For pose evaluation, we report the RMSE of the absolute pose error for rotation
(APE$_R$, degrees) and translation (APE$_t$, metres), along with the median
absolute error of the $y$-axis (vertical) translation component ($\delta_y$,
metres). For structure quality, we use the structure metrics from Section~\ref{sec:c2g_expt}

\smallskip
\noindent \textbf{Observation 1:} \textit{G3T-Long has lower ambient frame pose errors.} From Table~\ref{tab:inc_recon_tum_pose}, we can see that, in most cases, performing incremental 3D reconstruction using \OurMethod-Long leads to better performance on APE$_R$ and APE$_t$. This suggests that performing 1-DoF rotation constrained incremental 3D reconstruction using gravity-aligned pointmaps is more robust. Additionally, we observe that \OurMethod generally has small $\delta_y$ values, unlike VGGT-Long which has high variance across scenes. Since the upward direction is aligned with the $y$-axis for \OurMethod, this implies that our gravity-aligned incremental reconstruction procedure reduces vertical drift.

\smallskip
\noindent \textbf{Observation 2:} \textit{G3T-Long has better structure metrics.} From Table~\ref{tab:inc_recon_tum_pose}, we can see that \OurMethod-Long has significant improvements over VGGT-Long on structure metrics. These results complement those observed in Table \ref{tab:inc_recon_tum_pose}, as they follow from superior chunk alignment achieved in the gravity frame.

\begin{figure}[t]
    \centering
    \includegraphics[width=\textwidth]{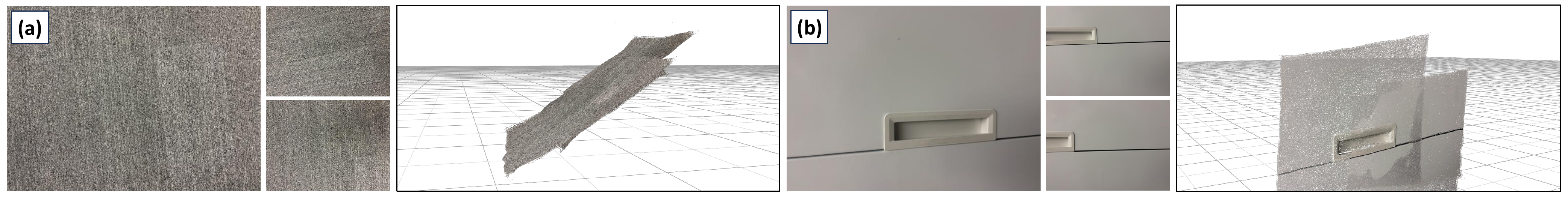}
    \vspace{-10pt}
    \caption{\textbf{Visualizing failure cases.} \OurMethod struggles to resolve upright alignment in scenes with ambiguous structural cues. We illustrate this with two self-captured examples. In (a), close-up floor images cause \OurMethod to produce a slanted pointmap. In (b), horizontally rotated images of a vertically-aligned cabinet lead to pointmaps with incorrect orientation.} 
    \label{fig:fail_case_viz}
    \vspace{-10pt}
\end{figure}

\section{Discussion}
\label{sec:disc}

Our work highlights an underexplored aspect of 3D pointmap prediction models---the choice of coordinate frame---and argues for the use frames with more structure, in particular \emph{gravity-aligned frames}. Beyond producing accurate upright-aware predictions, we demonstrate that we can perform robust submap-based incremental 3D reconstruction by leveraging the reduced degrees of freedom of gravity-aligned frames. While our work takes a step towards consideration of structured coordinate frames, we note some limitations and directions for future work. 

\smallskip
\noindent \textbf{Limitations.} \OurMethod may not produce good upright-aware predictions in scenes with ambiguous structural cues. For example, \OurMethod can struggle to estimate upright pointmaps from close-up images of floors and walls if additional unambiguous context is not present (see Figure~\ref{fig:fail_case_viz}).

\smallskip
\noindent \textbf{Future work.} There are many exciting directions for future work. For instance, it would be interesting to explore if gravity-aligned prediction naturally encourages performance improvement in tasks that favor uprightedness, such as physically based simulation or spatial reasoning. 
Additionally, it could also be advantageous to explore coordinate frames with even more structure, such as Manhattan frames (when appropriate), or even compositions of coordinate frames (e.g., a mixture of an overall scene coordinate frame, as well as local object coordinate frames for items in the scene).

\section*{Acknowledgements}
\label{sec:ack}

We thank the authors of DUSt3R~\cite{Wang_2024_CVPR}, VGGT~\cite{wang2025vggt}, CUT3R~\cite{cut3r}, and VGGT-Long~\cite{deng2025vggtlongchunkitloop} for open-sourcing their projects, which our work builds upon. Additionally, we would like to thank Aditya Chetan, Haian Jin and Jay Karhade for their feedback on initial drafts of the paper. This work was funded in part by the National Science Foundation (IIS-2211259 and IIS-2212084), and benefited from the NVIDIA Academic Grant Program for compute resources.

%
%
\bibliographystyle{plain}
\bibliography{main}


\newpage
\appendix

\section*{Supplementary Material}

\section{Additional qualitative results}
\label{sec:supp_additional_qual}

\begin{figure}[t]
    \centering
    \includegraphics[width=\textwidth]{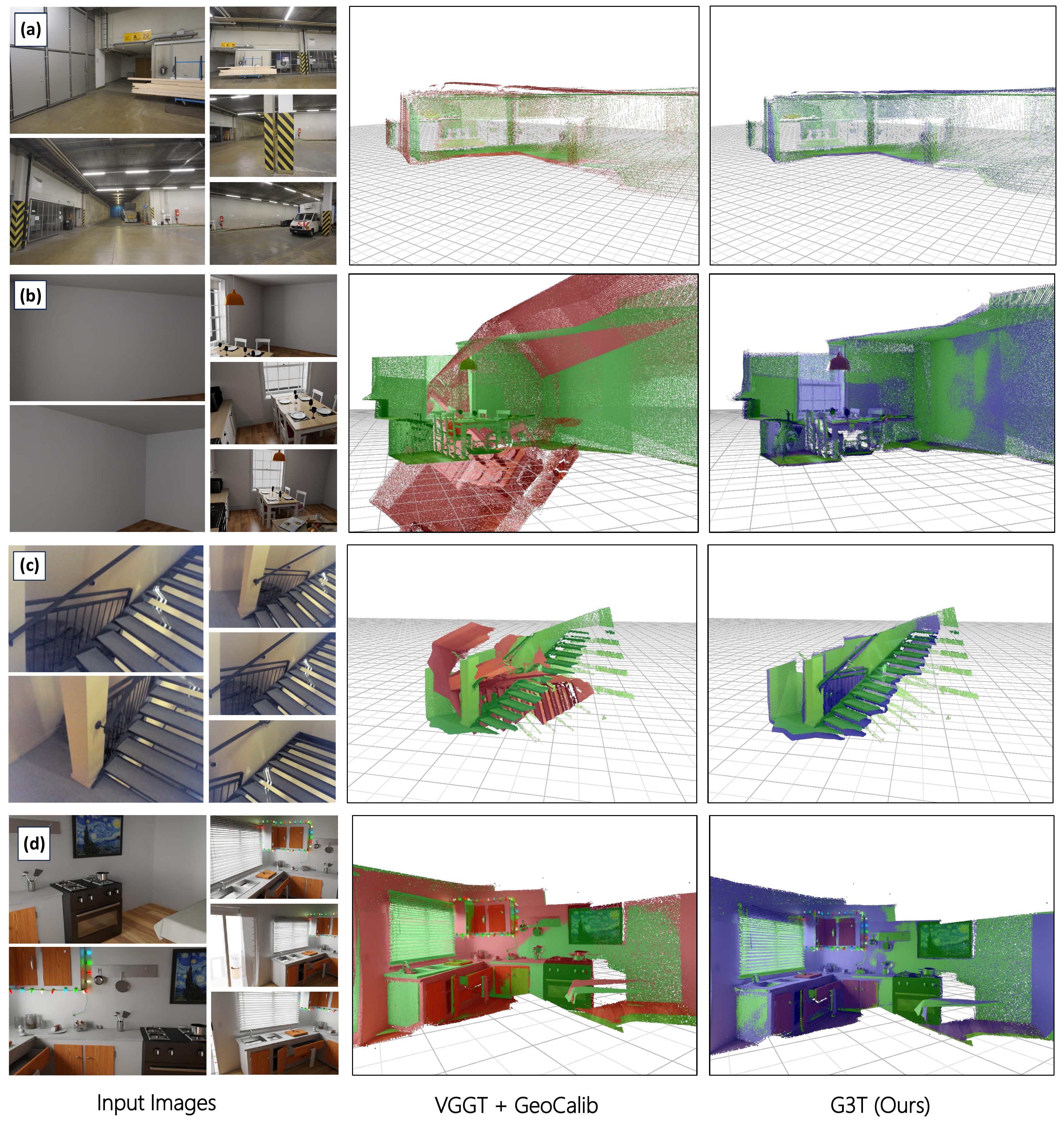}
    \caption{\textbf{\OurMethod can consistently place pointmaps in an upright frame.} For the given set of input images, we compare VGGT predictions (\textbf{\color{red}{red}}) made upright using GeoCalib, and our \OurMethod predictions (\textbf{\color{blue}{blue}}) with ground-truth gravity-aligned pointmaps (\textbf{\color{forestgreen}{green}}). We also render a grid depicting a plane parallel to the ground. We observe that the composition of VGGT and GeoCalib often does not produce sufficiently upright pointmap predictions. In contrast, \OurMethod produces pointmaps that are better aligned along the vertical direction.} 
    \label{fig:supp_mv}
\end{figure}

We have attached additional qualitative results in Figure \ref{fig:supp_mv}. We can see that \OurMethod achieves better gravity-aligned reconstructions than using the combination of VGGT and GeoCalib (similar to Figure~\ref{fig:mv_recon} in the main text), even in cases where the input images exhibit challenging roll and pitch angles.

\section{Dataset pre-processing details}
\label{sec:supp_dataset_details}

\smallskip
\noindent \textbf{Preparing data for training.} As described in Section~\ref{sec:training_details}, we use five large-scale datasets (MegaDepth~\cite{megadepth18}, Hypersim~\cite{hypersim21}, ARKitScenes~\cite{dehghan2021arkitscenes}, DL3DV~\cite{ling2024dl3dv} and TartanAir~\cite{tartanair2020iros}) for fine-tuning. We use the instructions and code from the \duster~\cite{Wang_2024_CVPR} codebase to download and pre-process data for the MegaDepth and ARKitScenes datasets. We use the instructions and code from the CUT3R~\cite{cut3r} codebase to download and pre-process data for the Hypersim, DL3DV and TartanAir datasets.

To fine-tune \OurMethod, we need the ground truth pointmaps to exist in the appropriate gravity-aligned coodinate frames (as defined in Section~\ref{sec:gravity_coords}). To that end, we require that each scene in each dataset has a world coordinate frame such that the $y$-axis is aligned with the up direction. This helps us calculate the rotations needed to transform points that exists in the camera frame to the gravity-aligned frame for a given image.

We noticed that the world coordinate frames for the scenes in the MegaDepth dataset are already gravity-aligned such that their $y$-axis is aligned with the up direction. However, for ARKitScenes, TartanAir and Hypersim, the world coordinate frame of the scenes uses the convention that their $z$-axis is aligned with the up direction. For these three datasets, we rotate the ground truth data such that the world coordinate frames have their $y$-axis aligned with the up direction. Finally, we noticed that the DL3DV dataset (downloaded via the CUT3R codebase) is not gravity-aligned. For this, we used the \texttt{model\_orientation\_aligner} module from COLMAP~\cite{schoenberger2016mvs, schoenberger2016sfm} to reorient the ground-truth poses such that the scenes lie in a world coordinate system which is gravity-aligned such that their $y$-axis is aligned with the up direction.

\smallskip
\noindent \textbf{Preparing data for evaluation.} For camera-to-gravity pose estimation and multi-view structure estimation experiments (Section~\ref{sec:c2g_expt}), we use three datasets (7Scenes~\cite{shotton2013scene}, NRGBD~\cite{Azinovic_2022_CVPR} and ETH3D~\cite{schoeps2017cvpr}). For submap-based incremental 3D reconstruction (Section~\ref{sec:incremental_3D_expt}), we choose sequences from the TUM RGBD dataset~\cite{sturm12benchmark} for evaluation. Do note that all these datasets were not seen by both VGGT and \OurMethod during training. 

We preprocess 7Scenes by following code and instructions from the Spann3R~\cite{wang20243dspann3r} and CUT3R~\cite{cut3r} codebases, NRGBD by adapting some code from the CUT3R codebase, and ETH3D by adapting some code from the Mp-SfM~\cite{pataki2025mpsfm} codebase. For preprocessing TUM RGBD sequences, we use code from the gradSLAM~\cite{gradslam} codebase to synchronize RGB, depth and pose data by timestamps. We then subsample the synchronized TUM RGBD data by selecting every 5th frame.

Similar to the training datasets, we also need to gravity-align the evaluation datasets. To this end, we use the \texttt{model\_orientation\_aligner} module from COLMAP~\cite{schoenberger2016mvs, schoenberger2016sfm} on all of the above datasets to make them gravity aligned (similar to how we processed the training datasets).

\begin{figure}[t]
    \centering
    \includegraphics[width=0.8\textwidth]{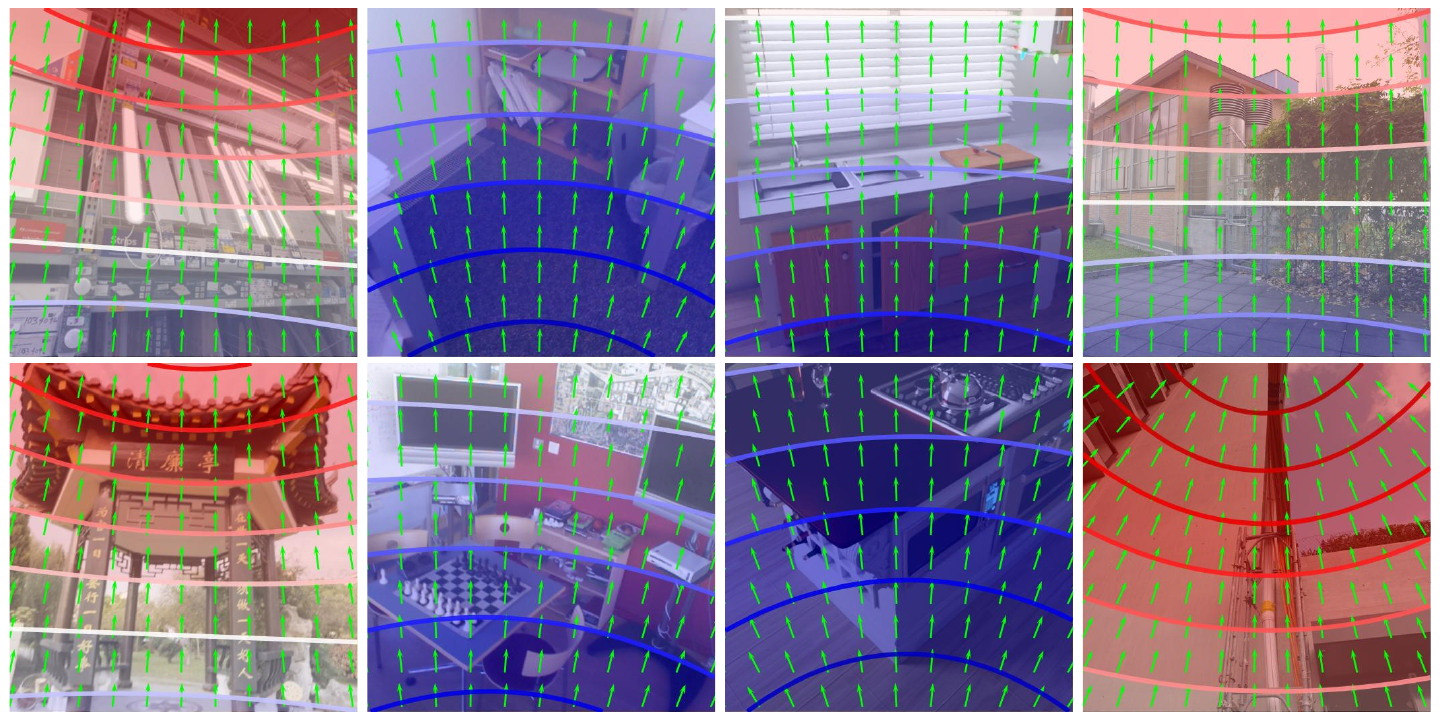}
    \caption{\textbf{Visualizing gravity-aligned ground-truth created using model\_orientation\_aligner.} We visualize perspective fields~\cite{jin2022PerspectiveFields} using gravity-aligned data created using \texttt{model\_orientation\_aligner} for a few samples from the DL3DV, 7Scenes, NRGBD and ETH3D datasets. For all samples, we can see that the up vectors are aligned with vertical structures present in the scene, thus validating the reliability of using \texttt{model\_orientation\_aligner} to create ground-truth gravity-aligned data.} 
    \label{fig:model_orientation_aligner_viz}
\end{figure}

\smallskip
\noindent \textbf{On using model\_orientation\_aligner.} Prior work has used the \texttt{model\_orientation\_aligner} module from COLMAP to align scenes to canonical orientations. For instance, GeoCalib~\cite{veicht2024geocalib} used it to gravity-align MegaDepth~\cite{megadepth18} data for evaluation, and MegaScenes~\cite{tung2024megascenes} dataset used it to make their scenes Manhattan aligned. We follow this precedent to use \texttt{model\_orientation\_aligner} to gravity-align our datasets as well. As a sanity check, Figure~\ref{fig:model_orientation_aligner_viz} visualizes perspective fields on gravity-aligned samples. We can see that up arrows are strongly aligned with the scene's vertical direction. We additionally evaluate on synthetic datasets that are naturally gravity-aligned (i.e., not processed by \texttt{model\_orientation\_aligner}) in Table~\ref{tab:c2g_hsim}. The trends observed here are consistent with those in Table~\ref{tab:c2g}. This suggests that data aligned with \texttt{model\_orientation\_aligner} is reliable for both training and evaluation.

\smallskip
\noindent \textbf{Sampling multi-view data.} To sample $N$ views from a scene in a dataset, we construct a scene graph data structure that stores the visibility relationships between nodes. For the MegaDepth and ARKitScenes datasets, we build scene graphs using the pair metadata from the \duster~codebase. For all other training and evaluation datasets, we construct complete scene graphs where every pair of nodes is connected by an edge.

During training, we sample $N$ views from a scene by first randomly selecting an initial node, then randomly selecting $N-1$ of its neighbors from the scene graph.

For the the multi-view pointmap structure experiment (Section~\ref{sec:c2g_expt}), we adopt a more controlled sampling strategy. We assign a distance value to each edge in the scene graph by summing the rotation difference (computed using Equation 5) and the translation difference (computed using the $L^2$ norm on mean-normalized translations). We then iteratively sample $N-1$ neighbor nodes for a randomly selected initial node such that all pairwise distances exceed 0.5, falling back to random neighbor sampling if this fails more than 100 times. This sampling strategy ensures adequate spatial separation between the sampled views.

For the camera-to-gravity pose estimation experiment (Section~\ref{sec:c2g_expt}), we evaluate performance with 1, 4, and 8 input views. We first sample 8 views following the same procedure as the multi-view pointmap prediction experiment, then construct the 1- and 4-view cases by taking the first 1 and 4 views of each 8-view sample. This ensures the cases are nested subsets, allowing us to cleanly measure the effect of adding more views.

\begin{table}[t!]
\centering
\caption{
\textbf{Camera to gravity rotation estimation results.} This table mirrors Table~\ref{tab:c2g} from the main paper, but performs evaluation on held out scenes from the Hypersim and TartanAir dataset. These are synthetic datasets which are already gravity aligned (i.e., these datasets were not gravity-aligned using \texttt{model\_orientation\_aligner}). The trends observed here are consistent with those in Table~\ref{tab:c2g}.
}
\label{tab:c2g_hsim}
\resizebox{\textwidth}{!}{%
\begin{tabular}{llccccccccccccccc}
\toprule
\multirow{3}{*}{Dataset} & \multirow{3}{*}{Model} & \multicolumn{5}{c}{1V} & \multicolumn{5}{c}{4V} & \multicolumn{5}{c}{8V} \\
\cmidrule(lr){3-7} \cmidrule(lr){8-12} \cmidrule(lr){13-17}
 & & \multicolumn{2}{c}{$R_{\text{err}}$ ($\downarrow$)} & \multicolumn{3}{c}{$R_{\text{acc}}$ ($\uparrow$)} & \multicolumn{2}{c}{$R_{\text{err}}$ ($\downarrow$)} & \multicolumn{3}{c}{$R_{\text{acc}}$ ($\uparrow$)} & \multicolumn{2}{c}{$R_{\text{err}}$ ($\downarrow$)} & \multicolumn{3}{c}{$R_{\text{acc}}$ ($\uparrow$)} \\
\cmidrule(lr){3-4} \cmidrule(lr){5-7} \cmidrule(lr){8-9} \cmidrule(lr){10-12} \cmidrule(lr){13-14} \cmidrule(lr){15-17}
 & & Mean & Median & @1\textdegree & @2\textdegree & @5\textdegree & Mean & Median & @1\textdegree & @2\textdegree & @5\textdegree & Mean & Median & @1\textdegree & @2\textdegree & @5\textdegree \\
\midrule
\multirow{3}{*}{Hypersim} & GeoCalib & 3.69 & 1.88 & 34.40 & 52.40 & 89.60 & 3.77 & 1.79 & 35.20 & 54.00 & 88.00 & 3.72 & 1.73 & 36.40 & 55.20 & 88.80 \\
& Procrustes & \nextBestMethod{1.49} & \nextBestMethod{0.38} & \nextBestMethod{87.60} & \bestMethod{95.20} & \nextBestMethod{95.60} & \nextBestMethod{0.98} & \nextBestMethod{0.41} & \nextBestMethod{80.40} & \nextBestMethod{92.40} & \nextBestMethod{96.80} & \nextBestMethod{0.91} & \nextBestMethod{0.34} & \nextBestMethod{86.00} & \nextBestMethod{93.60} & \nextBestMethod{97.20} \\
& Local Head & \bestMethod{1.34} & \bestMethod{0.22} & \bestMethod{91.20} & \nextBestMethod{95.20} & \bestMethod{96.00} & \bestMethod{0.58} & \bestMethod{0.17} & \bestMethod{94.00} & \bestMethod{96.00} & \bestMethod{98.00} & \bestMethod{0.59} & \bestMethod{0.17} & \bestMethod{94.00} & \bestMethod{95.60} & \bestMethod{98.40} \\
\midrule
\multirow{3}{*}{TartanAir} & GeoCalib & 5.68 & 2.91 & 19.20 & 40.00 & 67.60 & 5.76 & 2.94 & 22.80 & 39.60 & 66.80 & 5.92 & 2.73 & 25.60 & 41.60 & 65.60 \\
& Procrustes & \bestMethod{1.49} & \nextBestMethod{0.89} & \nextBestMethod{54.00} & \bestMethod{76.00} & \bestMethod{95.60} & \nextBestMethod{1.74} & \nextBestMethod{1.19} & \nextBestMethod{44.80} & \nextBestMethod{71.20} & \nextBestMethod{94.80} & \nextBestMethod{1.67} & \nextBestMethod{1.09} & \nextBestMethod{47.60} & \nextBestMethod{71.60} & \nextBestMethod{96.00} \\
& Local Head & \nextBestMethod{1.52} & \bestMethod{0.82} & \bestMethod{56.40} & \nextBestMethod{74.80} & \nextBestMethod{94.80} & \bestMethod{1.48} & \bestMethod{0.99} & \bestMethod{50.80} & \bestMethod{76.80} & \bestMethod{96.80} & \bestMethod{1.41} & \bestMethod{0.92} & \bestMethod{51.60} & \bestMethod{77.20} & \bestMethod{97.20} \\
\bottomrule
\end{tabular}%
}
\end{table}

\section{Additional implementation details}
\label{sec:supp_additional_impl}

In this section we elaborate on implementation details. Note that we also plan to open-source our codebase.

\smallskip
\noindent \textbf{Local and relative camera heads.} In Section~\ref{sec:gravity_coords}, we briefly introduced the local and relative camera heads in \OurMethod. Here, we provide additional implementation details.

Both heads are derived from the original VGGT camera head implementation with minor modifications. Recall that the original camera head outputs parameters ${\mathcal{G} = \{G_i\}}$, where $G_i \in \mathbb{R}^9$ (as explained in Section 3.1). In contrast, the local camera head of \OurMethod outputs parameters $\mathcal{G}^l = \{G^l_i\}$ where $G^l_i \in \mathbb{R}^6$, and the relative camera head of \OurMethod outputs parameters $\mathcal{G}^r = \{G^r_i\}$ where $G^r_i \in \mathbb{R}^5$. To accommodate the reduced output dimensionality, we shrink the empty pose token parameter size and the widths of a few linear layers accordingly, then initialize all weights from the pre-trained original camera head.

Additionally, in Section~\ref{sec:gravity_coords}, we mentioned that the relative camera head output $\mathcal{G}^r = \{G^r_i\}$ contains $q^r_i \in \mathbb{R}^2$ which represents a 1-DoF relative yaw w.r.t.\ the first frame. Here, $q^r_i$ cpatures the $y$ and $w$ components of a quaternion, as the $x$ and $z$ components are 0 for a pure 1-DoF rotation along the $y$-axis.

\begin{table}[t!]
\centering
\caption{
\textbf{Submap-based incremental 3D reconstruction on TUM RGBD: Ambient frame pose metrics (Depthmap unprojection).} This table mirrors Table~\ref{tab:inc_recon_tum_pose} from the main paper, but uses pointmaps obtained by unprojecting depthmaps with estimated poses. The trends are consistent with those in Table~\ref{tab:inc_recon_tum_pose}.
}
\label{tab:inc_recon_tum_pose_D}
\resizebox{\linewidth}{!}{%
\setlength{\tabcolsep}{4pt}
\begin{tabular}{llcccccccccc}
\toprule
Method & Metric ($\downarrow$) & fr1/360 & fr1/desk & fr1/desk2 & fr1/room & fr1/plant & fr1/teddy & fr3/loh\footnotemark & fr2/ps\footnotemark[\value{footnote}] & fr2/ps2\footnotemark[\value{footnote}] & fr2/ps3\footnotemark[\value{footnote}] \\
\midrule
\multirow{3}{*}{VGGT$_D$} & APE$_R$ & \bestMethod{16.790} & 2.247 & \bestMethod{1.676} & 3.962 & 3.247 & 4.507 & \bestMethod{2.844} & 5.447 & 8.970 & 6.775 \\
& APE$_t$ & 0.051 & 0.021 & 0.037 & 0.179 & 0.051 & 0.087 & 0.226 & 0.415 & 0.398 & 0.980 \\
& $\delta_y$ & 0.022 & \bestMethod{0.004} & \bestMethod{0.006} & 0.046 & 0.020 & 0.036 & 0.105 & 0.209 & 0.194 & \bestMethod{0.043} \\
\midrule
\multirow{3}{*}{\OurMethodD} & APE$_R$ & 18.429 & \bestMethod{0.735} & 4.527 & \bestMethod{2.502} & \bestMethod{1.740} & \bestMethod{1.777} & 4.127 & \bestMethod{2.776} & \bestMethod{3.339} & \bestMethod{6.329} \\
& APE$_t$ & \bestMethod{0.049} & \bestMethod{0.015} & \bestMethod{0.031} & \bestMethod{0.150} & \bestMethod{0.035} & \bestMethod{0.057} & \bestMethod{0.178} & \bestMethod{0.286} & \bestMethod{0.232} & \bestMethod{0.211} \\
& $\delta_y$ & \bestMethod{0.010} & 0.005 & 0.014 & \bestMethod{0.026} & \bestMethod{0.018} & \bestMethod{0.017} & \bestMethod{0.022} & \bestMethod{0.028} & \bestMethod{0.032} & \bestMethod{0.043} \\
\bottomrule
\end{tabular}%
}
\end{table}
\begin{table}[t!]
\centering
\caption{
\textbf{Submap-based incremental 3D reconstruction on TUM RGBD: Structure metrics (Depthmap unprojection).} This table mirrors Table~\ref{tab:inc_recon_tum_struc} from the main paper, but uses pointmaps obtained by unprojecting depthmaps with estimated poses. The trends are consistent with those in Table~\ref{tab:inc_recon_tum_struc}.
}
\label{tab:inc_recon_tum_struc_D}
\resizebox{\linewidth}{!}{%
\setlength{\tabcolsep}{4pt}
\begin{tabular}{llcccccccccc}
\toprule
Method & Metric & fr1/360 & fr1/desk & fr1/desk2 & fr1/room & fr1/plant & fr1/teddy & fr3/loh\footnotemark[\value{footnote}] & fr2/ps\footnotemark[\value{footnote}] & fr2/ps2\footnotemark[\value{footnote}] & fr2/ps3\footnotemark[\value{footnote}] \\
\midrule
\multirow{3}{*}{VGGT$_D$} & ACC ($\downarrow$) & 0.053 & \bestMethod{0.009} & \bestMethod{0.014} & 0.051 & 0.032 & 0.029 & 0.047 & 0.102 & 0.147 & 0.447 \\
& COMP ($\downarrow$) & 0.065 & \bestMethod{0.011} & \bestMethod{0.014} & 0.047 & \bestMethod{0.036} & 0.040 & 0.034 & 0.045 & 0.108 & 0.382 \\
& NC ($\uparrow$) & \bestMethod{0.681} & \bestMethod{0.616} & \bestMethod{0.597} & \bestMethod{0.641} & \bestMethod{0.607} & 0.588 & 0.612 & 0.670 & 0.704 & 0.581 \\
\midrule
\multirow{3}{*}{\OurMethodD} & ACC ($\downarrow$) & \bestMethod{0.042} & 0.010 & \bestMethod{0.014} & \bestMethod{0.041} & \bestMethod{0.026} & \bestMethod{0.022} & \bestMethod{0.036} & \bestMethod{0.063} & \bestMethod{0.073} & \bestMethod{0.084} \\
& COMP ($\downarrow$) & \bestMethod{0.056} & \bestMethod{0.011} & \bestMethod{0.014} & \bestMethod{0.039} & 0.042 & \bestMethod{0.038} & \bestMethod{0.028} & \bestMethod{0.038} & \bestMethod{0.068} & \bestMethod{0.067} \\
& NC ($\uparrow$) & 0.692 & 0.608 & 0.595 & 0.661 & 0.615 & \bestMethod{0.599} & \bestMethod{0.632} & \bestMethod{0.714} & \bestMethod{0.720} & \bestMethod{0.740} \\
\bottomrule
\end{tabular}%
}
\end{table}

\smallskip
\noindent \textbf{VGGT-Long global optimization.} In Section~\ref{sec:gravity-aligned-procrustes}, we described the global optimization procedure of VGGT-Long (Equation 2) and our gravity-aligned variant (Equation 4), noting that the norms in both equations are applied to Lie algebra representations of $\delta_\pi$ and $\delta_{\pi^y}$, respectively. Here we provide further details.

The $\simthreelie$ Lie algebra is the 7-dimensional tangent space of $\simthree$, with 3 translation, 3 rotation, and 1 scale components. The logarithmic and exponential maps convert between $\simthree$ and $\simthreelie$ representations. The VGGT-Long~\cite{deng2025vggtlongchunkitloop} codebase uses the PyPose~\cite{wang2023pypose} library for these operations, and defines the global optimization objective of Equation 2 in $\simthreelie$ space. This is advantageous because elements of $\simthreelie$ are unconstrained, allowing optimization to proceed without constraints.

For our gravity-aligned variant, we instead work with $\simthreeGA$ poses and their corresponding 5-dimensional Lie algebra $\simthreeGAlie$. Since $\simthreeGA$ restricts rotations to the $y$-axis, the rotation component of $\simthreeGAlie$ is 1-dimensional. We modify the VGGT-Long codebase to support the gravity-aligned global optimization of Equation 4.

\smallskip
\noindent \textbf{Configuring submap-based incremental reconstruction.} Here, we discuss the hyperparameters used for the submap-based incremental 3D reconstruction experiments in Section~\ref{sec:incremental_3D_expt}. As described in Section~\ref{sec:gravity-aligned-procrustes}, VGGT-Long processes long sequences by breaking them into overlapping chunks. We set the maximum chunk size to 25 frames, with 7 overlapping frames between adjacent chunks. VGGT-Long also forms chunks across loop closure detections for use in the optimization process, controlled by a loop chunk size hyperparameter; in general, each loop closure chunk spans twice this value. We set the loop chunk size to 3.

\begin{figure}[t!]
    \centering
    \includegraphics[width=0.9\textwidth]{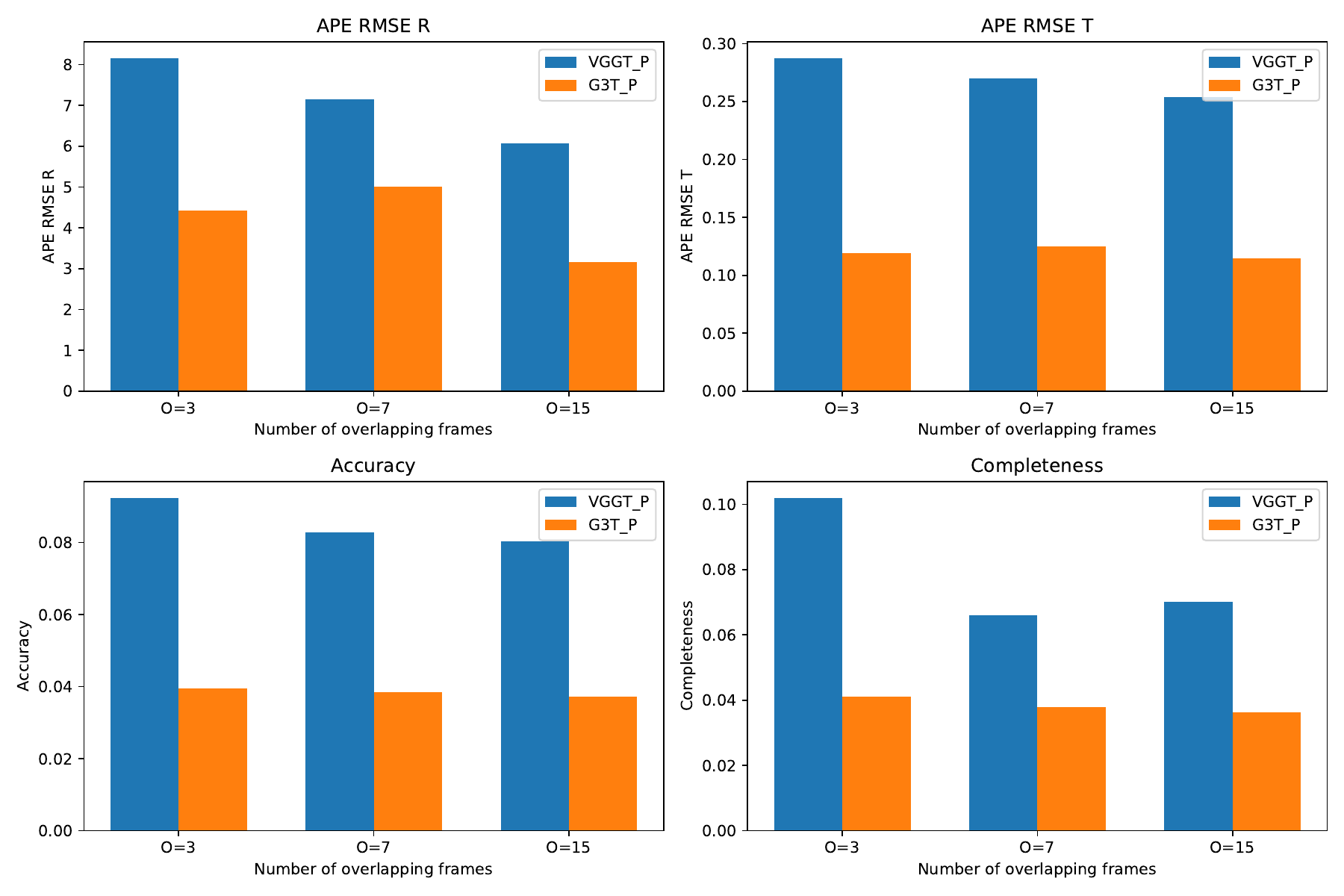}
    \caption{\textbf{Effect of number of overlapping frames in submap-based incremental reconstruction.} Here, we plot pose metrics (APE$_R$, APE$_t$) and structure metrics (ACC, COMP) as a function of number of overlapping frames, averaged across the 10 sequences from Tables~\ref{tab:inc_recon_tum_pose} and~\ref{tab:inc_recon_tum_struc} (lower is better for all metrics). While performance generally improves with more overlapping frames, we observe that \OurMethod has a consistent lead across all settings, with the largest margin when the number of overlapping frames is small.} 
    \label{fig:supp_num_overlap}
\end{figure}

We additionally measure the effect of number of overlapping frames, as more overlap can intuitively lead to more stable chunk alignment estimates. Specifically, we repeat the experiments of Section~\ref{sec:incremental_3D_expt} with 3 and 15 overlapping frames (instead of 7), keeping all other hyperparameters fixed. Figure~\ref{fig:supp_num_overlap} plots pose metrics (APE$_R$, APE$_t$) and structure metrics (ACC, COMP) as a function of number of overlapping frames, averaged across the 10 sequences from Tables~\ref{tab:inc_recon_tum_pose} and~\ref{tab:inc_recon_tum_struc} (lower is better for the plotted metrics). Performance generally improves with more overlapping frames. Nevertheless, \OurMethodP maintains a consistent lead across all settings, with the largest margin when the number of overlapping frames is small.

\smallskip
\noindent \textbf{GA-Procrustes.} For Procrustes, we use the \texttt{rigid\_points\_registration} function from the open-source RoMA library~\cite{bregier2021deepregression}. For GA-Procrustes, we implement a modified version incorporating the changes described in Section~\ref{sec:gravity-aligned-procrustes}.

\footnotetext{We shorten \texttt{long\_office\_household} to \texttt{loh} and \texttt{pioneer\_slam} to \texttt{ps} for brevity.}

\smallskip
\noindent \textbf{Camera-to-gravity pose estimation using Procrustes.} In Section~\ref{sec:c2g_expt}, we briefly introduced our algorithm to compute the camera-to-gravity pose given predictions from \OurMethod and VGGT using Procrustes alignment within a RANSAC loop. Here, we elaborate on the implementation details.

Given a set of $N$ images $\mathcal{I}$, we predict a set of gravity-aligned pointmaps $\mathcal{X}^{G_1}$ using \OurMethod and a set of camera-aligned pointmaps $\mathcal{X}^{C_1}$ using VGGT, along with their associated confidence maps $\Psi^{G_1}$ and $\Psi^{C_1}$. To prepare these predictions for robust alignment, we first pre-process the data as follows:

We create a joint confidence map $\Psi$ by element-wise multiplying the logarithms of the confidence maps from each model: $\Psi = \log \Psi^{G_1} \odot \log \Psi^{C_1}$. This ensures that regions with high confidence in both models have higher $\Psi$ values, while regions with low confidence in either model have lower $\Psi$ values. We then select the top 50\% of points based on their confidence values and independently scale-normalize them, where the scale is defined as the median distance of points to the pointmap center (the median of all points). We denote the resultant pointmaps as $\hat{\mathcal{X}}^{G_1}$ and $\hat{\mathcal{X}}^{C_1}$.

We then run a RANSAC loop for 5,000 iterations. In each iteration, we randomly sample 50,000 corresponding points from $\hat{\mathcal{X}}^{G_1}$ and $\hat{\mathcal{X}}^{C_1}$, and use the \texttt{rigid\_vectors\_registration} function from the open-source RoMA library~\cite{bregier2021deepregression} to estimate the rotation $R$ and scale $s$ (we discard the scale). We apply the predicted rotation matrix to transform $\hat{\mathcal{X}}^{C_1}$ and compare it with $\hat{\mathcal{X}}^{G_1}$ to count the number of inliers (defined as points with $L^2$ error less than 0.05). Finally, we return the rotation matrix $R$ with the highest number of inliers.

\smallskip
\noindent \textbf{On using GeoCalib.} Recall that VGGT outputs pointmaps in a camera coordinate frame. To create gravity-aligned pointmaps in this case, we use GeoCalib~\cite{veicht2024geocalib} to compute the camera to gravity transform. While GeoCalib estimates gravity from a single image, it also supports joint optimization across multiple images. For all experiments involving GeoCalib, we use the multi-image optimization mode.

\smallskip
\noindent \textbf{Pointmap visualizations.} Several figures in this paper compare the upright alignment of gravity-aligned predicted pointmaps against ground-truth pointmaps. For each visualization, we use pointmaps extracted directly from the point head (i.e., VGGT$_P$ and \OurMethodP). Since VGGT does not guarantee upright outputs, we apply GeoCalib with multi-image optimization to make its pointmaps gravity-aligned. We then apply GA-Procrustes to align each predicted pointmap to the ground-truth, producing a visual comparison that highlights deviations from uprightness.


\end{document}